\documentclass[conference]{IEEEtran}
\IEEEoverridecommandlockouts
\renewcommand\IEEEkeywordsname{Keywords}
\usepackage{cite}
\usepackage{amsmath,amssymb,amsfonts}
\usepackage{algorithmic}
\usepackage{graphicx}
\usepackage{textcomp}
\usepackage{xcolor}
\usepackage{hyperref}
\def\BibTeX{{\rm B\kern-.05em{\sc i\kern-.025em b}\kern-.08em
    T\kern-.1667em\lower.7ex\hbox{E}\kern-.125emX}}
\begin{document}

\title{Kinova Gemini: Interactive Robot Grasping with Visual Reasoning and Conversational AI}

\author{Hanxiao Chen, Jiankun Wang, \emph{Senior Member, IEEE} and Max Q.-H. Meng, \emph{Fwllow, IEEE} 
\thanks{This work is partially supported by Shenzhen Key Laboratory of Robotics Perception and Intelligence (ZDSYS20200810171800001), Southern University of Science and Technology, Shenzhen 518055, China, and National Natural Science Foundation of China grant \#62103181, \emph{(Corresponding authors:Jiankun Wang, Max Q.-H. Meng).}}
\thanks{Hanxiao Chen (2022 Master Student in Harbin Institute of Technology) and Jiankun Wang are with Shenzhen Key Laboratory of Robotics Perception and Intelligence, and the Department of Electronic and Electrical Engineering, Southern University of Science and Technology, Shenzhen 518055, China. {\tt\small e-mail: wangjk@sustech.edu.cn}}%
\thanks{Max Q.-H. Meng is with Shenzhen Key Laboratory of Robotics Perception and Intelligence, and the Department of Electronic and Electrical Engineering, Southern University of Science and Technology, Shenzhen 518055, China, on leave from the Department of Electronic Engineering, The Chinese University of Hong Kong, Hong Kong, and also with the Shenzhen Research Institute, The Chinese University of Hong Kong in Shenzhen, Shenzhen 518057, China. {\tt\small e-mail: max.meng@ieee.org}}%
}

\maketitle

\begin{abstract}
To facilitate recent advances in robotics and AI for delicate collaboration between humans and machines, we propose the Kinova Gemini, an original robotic system that integrates conversational AI dialogue and visual reasoning to make the Kinova Gen3 lite robot help people retrieve objects or complete perception-based pick-and-place tasks. When a person walks up to Kinova Gen3 lite, our Kinova Gemini is able to fulfill the user's requests in three different applications: (1) It can start a natural dialogue with people to interact and assist humans to retrieve objects and hand them to the user one by one. (2) It detects diverse objects with YOLO v3 and recognize color attributes of the item to ask people if they want to grasp it via the dialogue or enable the user to choose which specific one is required. (3) It applies YOLO v3 to recognize multiple objects and let you choose two items for perception-based pick-and-place tasks such as “Put the banana into the bowl” with visual reasoning and conversational interaction.
\end{abstract}

\begin{IEEEkeywords}
robot grasping, object detection, visual reasoning, human robot interaction, conversational AI
\end{IEEEkeywords}

\section{Introduction}
Robotic manipulation \cite{1} has attracted worldwide attention to develop more intelligent applications that could directly interact with the surrounding world and achieve different goals. The last decade has observed substantial growth in robotic manipulation research \cite{2,3,4,5,6}, which intends to exploit the increasing availability of affordable robot arms and grippers to invent robots for fundamental tasks (e.g., auto-grasping, door opening, object relocation), then such autonomous robots can be deployed for immense applications in hospitals, factories, restaurants, outer space and in countless others.

Most recently, Human-Centered AI (HCAI) has been proposed and considered as an emerging discipline intent on creating AI systems that amplify and augment rather than displace human abilities \cite{7}. Since HCAI seeks to preserve human control in a way that ensures AI to meet human needs while operating transparently, it has been rigorously investigated to design new forms of three important themes which are critical for the success of solid HCAI systems: (a) Human Robot Interaction (HRI) \& Human-AI collaboration or co-creation; (b) Responsible and human-compatible AI; (c) Natural language interaction.

Inspired by such HCAI elements, we develop the “Kinova Gemini” robotic system, where the 6-DoF Kinova Gen3 lite can not only automatically grasp multiple objects, but also integrates visual reasoning and natural language interaction to enhance human perception and carry out a dialogue with people to clarify their questions and fulfill the requests in different scenarios. 

Our article is structured as follows. First, we investigate related research work that runs through robot grasping detection and human robot interaction in Sec. II. Section III provides more detailed approach descriptions for robot perception, grasping execution, and dialogue interaction. Next, we discuss the “Kinova Gemini” HRI experiments and applications while exploring 3 different user scenarios in Sec. IV. Finally, Section V concludes our contributions, experiment results, and proposes future work to enhance “Kinova Gemini”.

\section{Related Work}

\subsection{Grasp Detection Methods}
\begin{figure*}[t]
\centering
\includegraphics[width=1.0\textwidth]{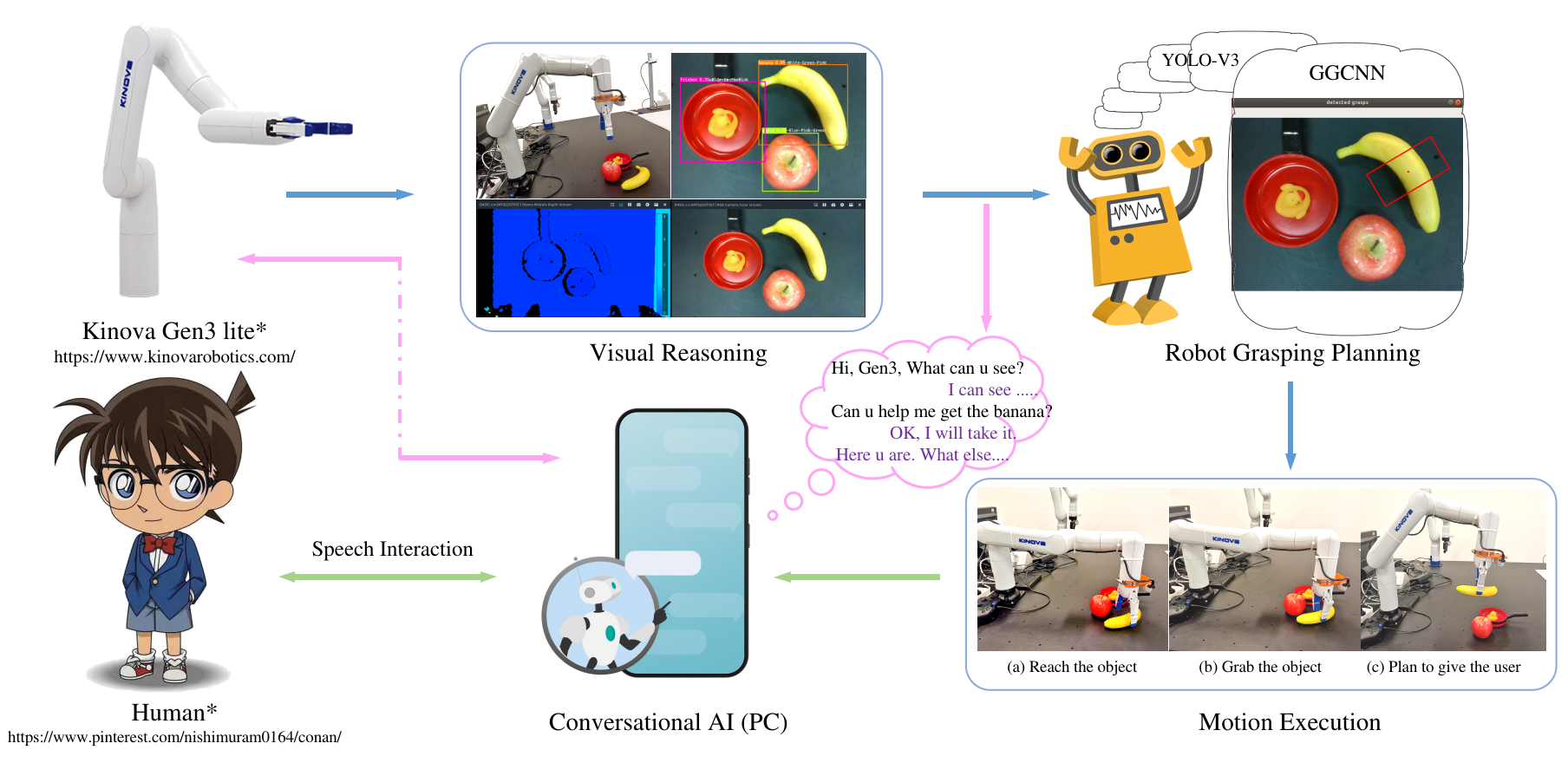} 
\caption{The proposed “Kinova Gemini” framework.}
\label{fig1}
\end{figure*}
Robot grasping has been studied extensively as the most basic manipulation task and a lot of grasping detection approaches \cite{8, 9, 10, 11} have been proposed to directly localize robotic grasping configurations from input noisy and partially occluded RGB-D images or point clouds, without assuming a known object CAD model. Actually, the existing various grasping detection approaches can achieve grasping success rates between 75\% and 95\% for novel objects presented in isolation or in light clutter. To begin with, Grasp Quality Convolutional Neural Networks (GQ-CNNs) \cite{8, 9} propose neural network architectures that take as input a depth image for grasping, then output the predicted probability that the gripper will successfully hold the item while lifting, transporting, and shaking the object. GQ-CNNs may be useful for quickly planning grasps that can pick \& place a variety of items on physical robots.

In addition, to facilitate the Dex-Net robot grasping project, Dex-Net 2.0 \cite{8} is designed to self-generate training datasets to learn GQ-CNN models that predict the success rate of candidate parallel-jaw grasps on objects via point clouds. Even the latest version of Dex-Net 4.0 \cite{9} unifies the reward metric across multiple grippers to train “ambidextrous” grasping policies that can determine which gripper is best for a particular object. Another classic grasping detection approach is Generative Grasping Convolutional
Neural Network (GG-CNN) \cite{10}, an ultra-lightweight, fully-convolutional network which predicts the pose and quality of antipodal grasps at every pixel in an input depth image. The lightweight and single-pass grasp generative nature of GG-CNN allows for fast execution and closed-loop control, which enables the accurate robot grasping in dynamic scenarios where objects are moving during grasping attempts. Grasp Pose Detection (GPD) \cite{11} has developed a package to detect 6-DOF grasping poses (3-DOF position and 3-DOF orientation) for the 2-finger robot hand (e.g., a parallel jaw gripper) in 3D point clouds. GPD takes a point cloud as input and produces pose estimates of viable grasps as output so that it can work for novel objects in dense clutter and enable more than just top-down grasps. 

\subsection{Human Robot Interaction}
As a field of study dedicated to understanding, designing, and evaluating robotic systems for use by or with humans, human-robot interaction (HRI) plays an important role in achieving much advanced HCAI systems or applications, especially in assistive robotics for healthcare. Martens et al. \cite{12} propose the semiautonomous robotic system “FRIEND” consisting of an electric wheelchair with the robot arm MANUS for people with upper limb impairments. Achic et al. \cite{13} develop an integral system combining a hybrid brain computer interface and shared control system for an electric wheelchair with an embedded robotic arm to help people achieve essential tasks like picking up a cup of water. In addition, \cite{14} explores effective design guidelines considering both graphical user interface (GUI) and tangible user interface (TUI) for HRI with assistive robot manipulation systems. 

In the digital era, speech-based interaction is gradually becoming the mainstream of HRI. Thus, Conversational AI \cite{15} has been proposed as the powerful HRI tool to lower the barriers of interaction, expand the user base, and demonstrate broad use cases and unlimited commercial values in various scenarios. For example, R Mead develop the “Semio” \cite{16}, a cloud-based platform that allows humans to use robots through natural communication-speech and body language. 

\section{Methods}
To successfully implement the “Kinova Gemini” robotic framework (Fig. 1) which enables the Kinova Gen3 lite to conduct visual reasoning to “see” the surrounding world, motion planning to grasp objects and conversational AI with humans to interact, we will discuss our applied approaches in the following three parts.

\subsection{Robot Perception}
After investigating diverse grasping detection methods in Sec. II that can achieve grasping success rates between 75\% and 95\% for multiple objects presented in isolation or in the light clutter, it is reasonable to apply the existing grasping policy detection approaches (e.g., GPD, Dex-Net, GG-CNN) for effective robot perception. Based on the practical grasping implementation rules, we investigate and successfully deploy the classical GG-CNN for grasping detection and YOLO v3 object detection for visual reasoning as follows.

\subsubsection{GG-CNN}
Unlike the previous deep-learning grasping techniques with discrete sampling of grasping candidates and long computation time, GG-CNN \cite{10} presents a real-time, object-independent grasping synthesis method for closed-loop grasping, which predicts the pose and quality of grasps at every pixel. In addition, the performance of GG-CNN has been evaluated via different scenarios by performing grasping trials with a Kinova Mico robot, achieving an 83\% grasp success rate on a set of previously unseen objects with adversarial geometry, 88\% on moving household objects, and 81\% in dynamic clutter. 

GG-CNN mainly considers the problem of detecting and executing antipodal grasps on unknown objects, perpendicular to a planar surface, given a depth image of the scene. In general, $g = (p, \phi, w, q)$ defines a grasp executed perpendicular to the $x$-$y$ plane, which is determined by its pose, i.e. the gripper's center position $p = (x, y, z)$ in Cartesian coordinates, the gripper's rotation $\phi$ around the $z$ axis and the required gripper width $w$. A scalar quality measure $q$, representing the chances of grasping success, is added to the pose. Since GG-CNN aims to detect grasps by a depth image $I = R^{H*W}$ taken from a camera with known intrinsic parameters, a grasp in the image $I$ can be described as $\tilde{g}=(s,\tilde{\phi},\tilde{w},q)$, where $s = (u, v)$ is the center point in image coordinates (pixels), $\tilde{\phi}$ is the rotation in the camera's reference frame and $\tilde{w}$ denotes the grasp width in image coordinates. 

A grasp in the image space $\tilde{g}$ can be converted to the world coordinates $g$ $(Eq. (1))$ by applying a sequence of transforms: (i) Hand eye calibration (Eye-in-hand) for $t_{RC}$ transformation from the camera frame to the robot frame; (ii) Camera intrinsic calibration for $t_{CI}$ transformation from 2D image coordinates to the 3D camera frame.
\begin{equation}
    g = t_{RC}(t_{CI}(\tilde{g}))
\end{equation}
Hence, the set of grasps in the image space can be referred to the grasping map, denoted as 
\begin{equation}
    G = (\Phi, W, Q) \in R^{3\times H \times W}
\end{equation}
where $\Phi$, $W$ and $Q$ $\in R^{H\times W}$ and include values of $\tilde{\phi}$, $\tilde{w}$ and $q$ respectively at each pixel $s$. Since GG-CNN directly calculates a grasp $\tilde{g}$ for each pixel in the depth image $I$, it defines a function $M$ from a depth image to the grasping map in the image coordinates: $ M (I) = G$. From $G$ we can calculate the best visible grasping in the image space $\tilde{g}^{*} = \max\limits_{Q} G$, and calculate the equivalent best grasping in world coordinates $g^{*}$ via $Eq. (1)$. 

\begin{figure}[t]
\centering
\includegraphics[width=1.0\columnwidth]{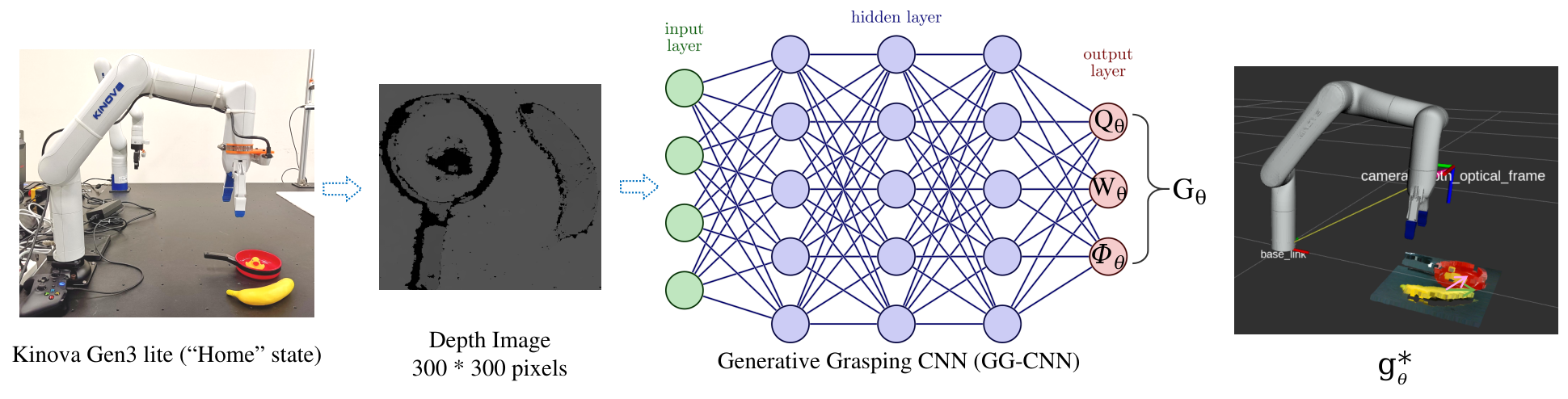} 
\caption{GG-CNN grasping detection by Kinova Gemini.}
\label{fig2}
\end{figure}

For GG-CNN training, $M_{\theta}$ denotes the neural network with $\theta$ being the network weights and $M_{\theta}(I) = (Q_{\theta},\Phi_{\theta}, W_{\theta}) \approx M(I)$  can be learned with a training set of inputs $I_{T}$ and corresponding outputs $G_{T}$ while applying the $L2$ loss function $L$, such that $\theta = \arg\min\limits_{\theta} L(G_{T}, M_{\theta}(I_{T}))$. According to \cite{10}, GG-CNN has been trained with a self-created dataset from the Cornell Grasping Dataset \cite{17} and the final network containing 62420 parameters can be saved for further evaluation on unseen objects, making it significantly smaller and faster to be applied on other robots for grasping detection in Fig. 2.

\subsubsection{YOLO v3}
You Only Look Once (YOLO) \cite{18} is a state-of-the-art, real-time object detection system that identifies specific objects in images, live feeds, or videos. Moreover, YOLO v3 is an improved version of YOLO and YOLO v2 to be implemented by Keras or OpenCV libraries and has been integrated into the robot grasping detection method to obtain the efficient visual information of surrounding environments. The main objective of YOLO v3 is applying features learned by a deep convolutional neural network to detect items. At first, it separates an image into grids and each grid cell can predict several number of boundary boxes (sometimes referred to as anchor boxes) around objects that score highly with the predefined aforementioned classes. Each boundary box has a respective confidence score of how accurate it assumes that prediction should be and detects only one object per bounding box. These boundary boxes are generated by clustering the dimensions of the ground truth boxes from the original datasets to find the most common shapes and sizes.

\begin{figure}[t]
\centering
\includegraphics[width=1.0\columnwidth]{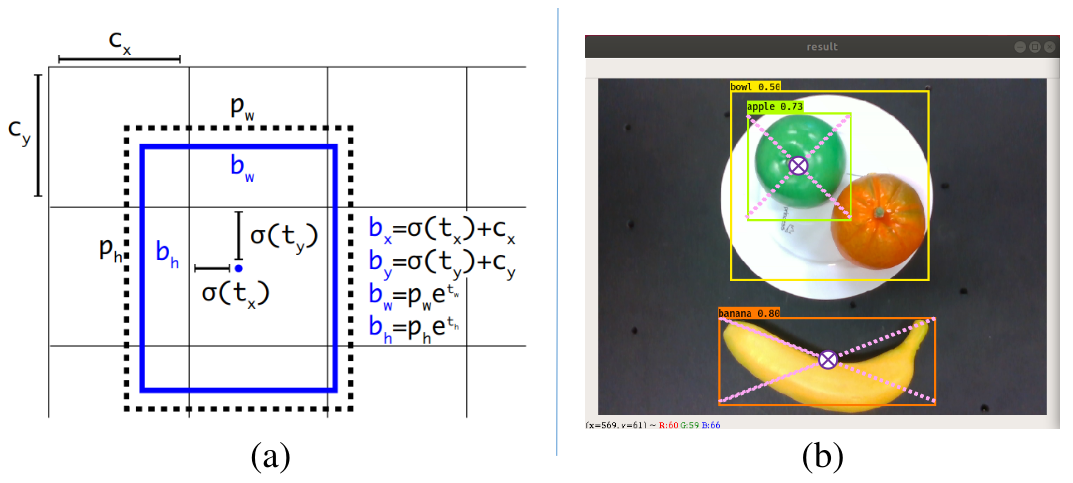}
\caption{(a) The generated bounding boxes with location prediction and dimension priors \cite{18}. (b) Real-time YOLO v3 object detection with the predicted bounding boxes.}
\label{fig2}
\end{figure}

Thanks to the open-source neural network architecture and trained YOLO v3 model weights, we can easily transfer the model in practical robot grasping scenarios for real-time object detection, which is significant for the robot to “see” the world and plan the following operations. As represented in Fig. 3, YOLO v3 predicts the width and height of the bounding box as offsets from cluster centriods and computes the center coordinates of the box relative to the location of filter application using a sigmoid function. In addition, the YOLO v3 object detection implemented by python often provides 4 vertex coordinates of the bounding box and we can computer the box center $(u, v)$ by easy addition and division in the image space $\tilde{g}$ . Inspired by \cite{19} that applies YOLO v3 in robot grasping detection, we consider it is reasonable to make the robot grasp objects within the planar surface by Eq. (1) transformations on the detected box center coordinates, which guides the Kinova Gen3 lite to reach objects exactly for successful grasping.

\subsection{Grasping Execution}
Based on the powerful Robot Operating System (ROS)\cite{20}, the official ROS Kortex package has been developed to interact with Kortex and Kinova robot products by building upon the KINOVA KORTEX API, which applies Google Protocol Buffer message objects to exchange data between client and server.

As we all know, MoveIt \cite{20} is the most advanced and flexible ROS library for motion planning and manipulation tasks, which integrates state-of-the-art path planning algorithms, inverse kinematics solvers, and collision detection techniques. In addition, ROS kortex package has auto-integrated MoveIt to configure the planning scene and send trajectories or pose goals to control the Kinova Gen3 lite robot with ROS nodes. Thus, we can directly achieve the effective grasping execution on the Kinova Gen3 lite robot by ROS Kortex and its wrapped MoveIt function.

\subsection{Dialogue Interaction}
HRI definitely requires the communication between robots and humans with diverse forms such as dialogue, vision, and action. Similar to human society, HRI with conversational AI dialogue can make the robotic system much natural and convenient to be integrated into human daily life. Combined with several well-developed libraries on speech recognition, text-to-speech (TTS) technique and natural language processing (NLP), our “Kinova Gemini” is strengthened to achieve dialogue interaction with humans.

Firstly, we install the “SpeechRecognition” python library to recognize human speech input from the USB microphone offline with CMU Sphinx and Google Cloud Speech API. Secondly, we can apply “pyttsx3” text-to-speech (TTS) conversation library to make the robot speak with human for interaction. Actually, “pyttsx3” can not only perform offline but also support multiple TTS engines including Sapi5, nsss, and espeak. Next, it's necessary to conduct deep analysis on the recognized speech to make robots capable of understanding and responding to human language. The industrial-strength “spaCy” NLP library features state-of-the-art speed, accuracy, and pretrained transformers like BERT for language tagging, parsing, named entity recognition, and text classification. To carry out a good dialogue with humans, Kinova Gemini can apply “spaCy” to clarify the user request and take advantage of voice information to finish complex tasks. 

\section{Experiments}
As the “Kinova Gemini” means, we especially develop our robotic system with a PC computer, an USB audio microphone, a 6 DoF Kinova Gen3 lite and an Intel RealSense D435i RGB-D camera, which is mounted to the robot wrist, approximately 90 mm above the fingertips and inclined at $10^\circ$ towards the gripper. The whole setup is shown in Fig. 4, and we primarily develop “Kinova Gemini” in three different cases and we will introduce them detailedly as follows.
\begin{figure}[t]
\centering
\includegraphics[width=1.0\columnwidth]{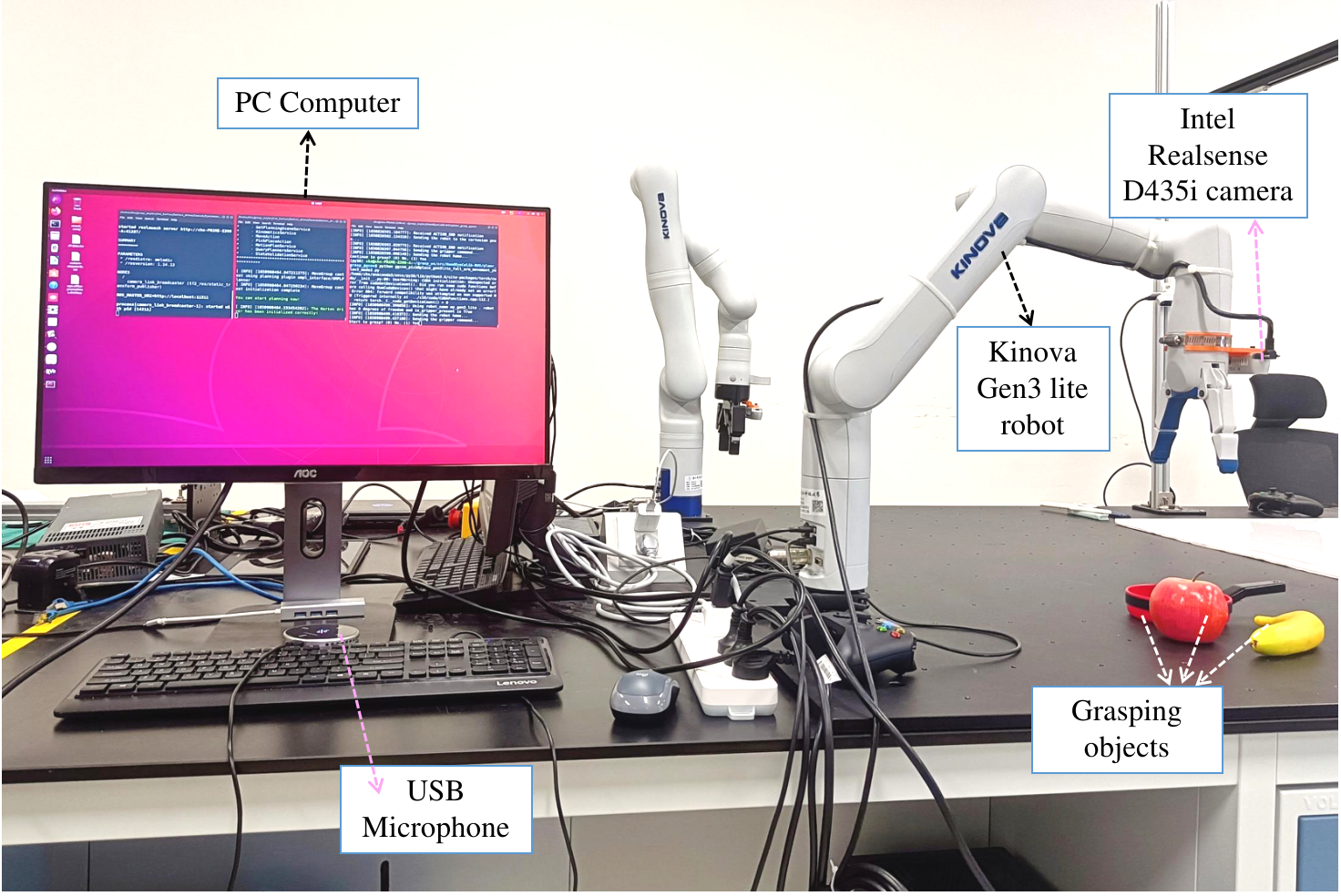}
\caption{The experiment setup with Kinova Gen3 lite.}
\label{fig4}
\end{figure}

\subsection{Mode 1 -- Talk to Retrieve Objects}
To begin with, we think that our “Kinova Gemini” should develop a natural dialogue with humans to interact and assist humans to retrieve diverse objects one by one and hand them to the user. According to Sec. III, robot perception plays a significant role in accurate grasping and we successfully implement GG-CNN for grasping detection on multiple unknown real-world objects (Fig. 5 (a)). Since GG-CNN can predict the quality and pose of grasps $G$ at every pixel by one-to-one mapping $M$ from a depth image $I$, we shall transform the center point $s = (u, v)$ of the detected grasp $g$ into the 3D world coordinates $W(x,y,z)$ by $Eq. 1$ with hand-eye calibration and camera intrinsic parameter calculation. Given the real-time predicted gripper width $W$, the grasping angle $\Phi$, and the grasping quality $Q$ from GG-CNN as shown in Fig. 5 (a), the Kinova Gemini can accurately reach the object to grasp it up and hand it to humans with the powerful ROS Kortex package. 
\begin{figure}[t]
\centering
\includegraphics[width=1.0\columnwidth]{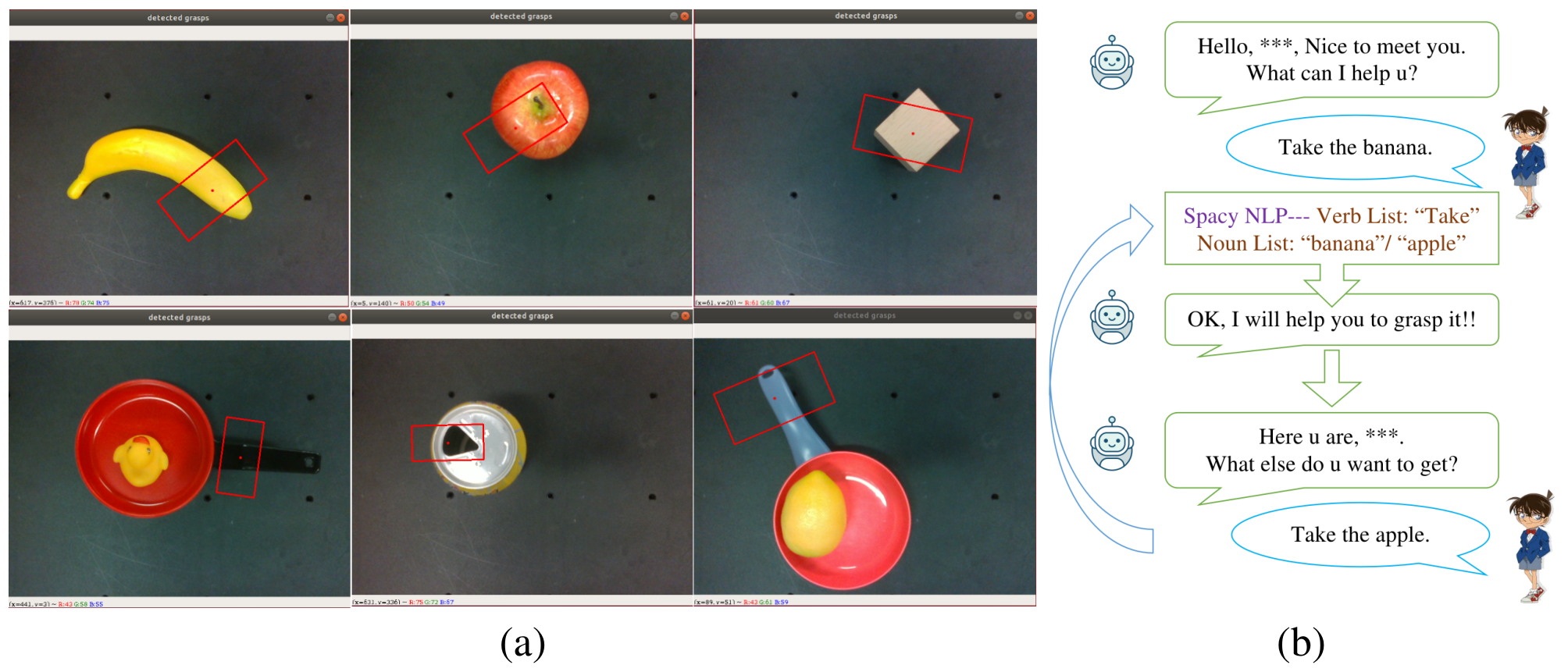}
\caption{(a) GG-CNN grasp detection results on multiple objects. (b) Human robot dialogue interaction in Mode 1.}
\label{fig5}
\end{figure}

In regard to dialogue interaction (Fig. 5 (b)), the Kinova Gemini will firstly greet with people to ask if they need some help. Then the user can speak out the short and clear request like “Take the banana” via the USB microphone. After receiving the voice information, our kinova will utilize the “spaCy” library to process and analyze the input sentence to extract verb and noun words for grasping matching, which means if the speech recognized “banana” is included in the pre-defined user-wanted object lists, the robot will say “Ok, I will help you to grasp it!!” and successfully grasp the object (Fig. 6) to humans, then respond with “Here you are. What else do you want to get?” to enter the following grasping looping steps for diverse unknown objects. More detailed successful experimental results can refer to \url{https://github.com/2000222/Kinova-Gemini}.

Generally speaking, “Kinova Gemini” in Mode 1 aims to help people fulfill their clear requests or commands with a natural and sustainable dialogue, which can be widely applied in practical human life for high-quality assistance and quick execution. To emphasize, GG-CNN performs exactly well with high quality scores for fast and accurate grasping detection on different objects including banana, apple, block, bottle, orange, spoon, bowl and so on, which lays the good foundation for MoveIt motion planning of ROS Kortex (Fig. 6). But in few times GG-CNN may return the depth of object as 0 because of the light change and object thickness. At that time, we can slightly adjust the setting position or angle of objects. Actually, this little disadvantage happens hardly and almost does not influence the combined framework of HRI dialogue interaction and grasping execution.

\begin{figure}[t]
\centering
\includegraphics[width=1.0\columnwidth]{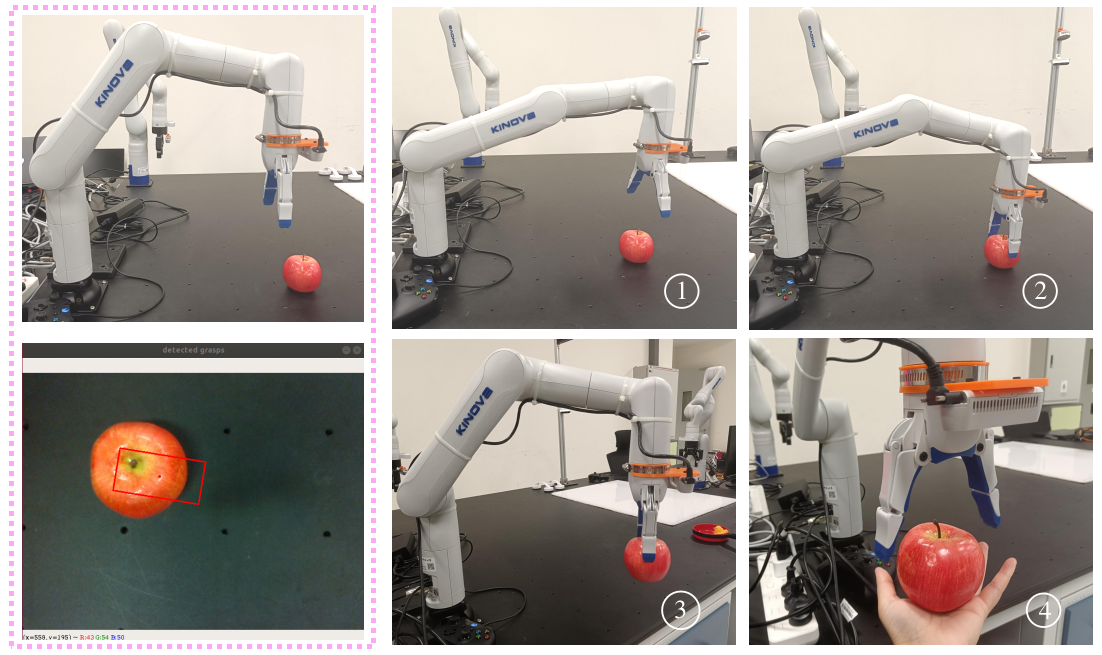}
\caption{Successful grasp-to-human steps with GG-CNN.}
\label{fig6}
\end{figure}

\subsection{Mode 2 -- Choose to Grasp Colorful Items}
In the second scenario, we enrich the “Kinova Gemini” with YOLO v3 object detection and color recognition for robot visual reasoning. Thanks to the easy usage of Kinova Gen3 lite and its flexible grippers, we successfully apply the YOLO v3 instead of GG-CNN for concrete visual perception with reasonable grasping width and angle values. Therefore, we can transform the bounding box center into the world coordinates and obtain the depth information of the detected center by pyrealsense library for effective graspings. Another essential function is color recognition that applies K-means clustering to predict prominent colors inside the detected object bounding box, which can benefit the user to completely understand object attributes. As shown in Fig. 7, the improved perception system of “Kinova Gemini” can return bounding boxes of multiple detected objects (e.g., apple, banana, bowl, spoon) with two or three dominant colors (e.g., Pink, Green, Black). In addition, we find sometimes the overlapping bounding boxes may influence the final predicted colors (Fig. 7 (c)) and the banana is often considered as “Green”.
\begin{figure}[t]
\centering
\includegraphics[width=1.0\columnwidth]{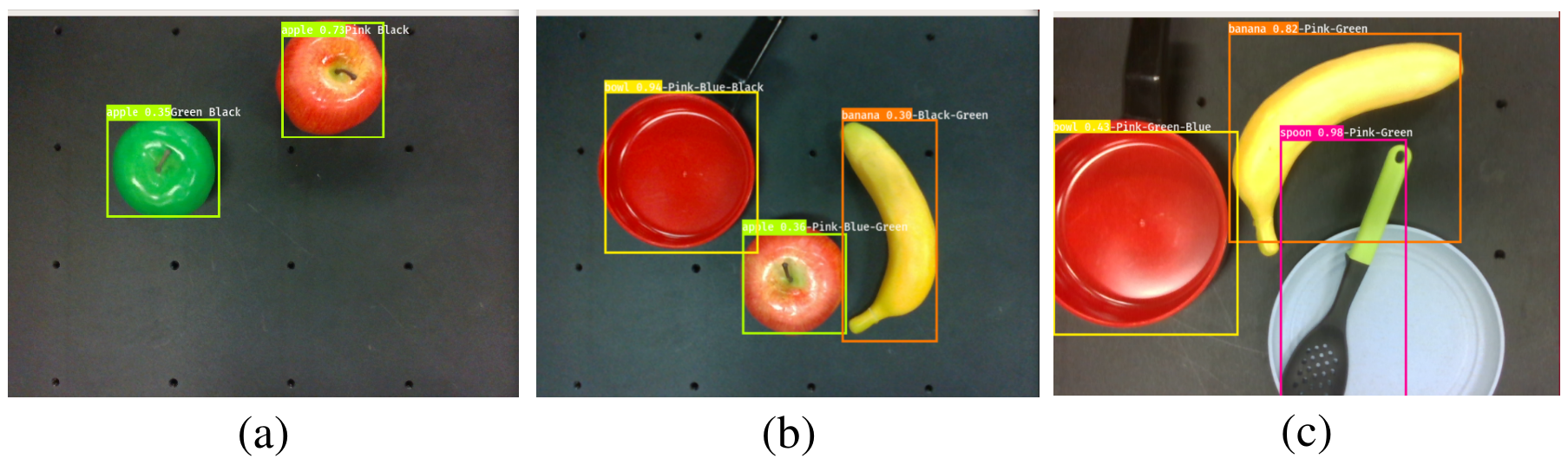}
\caption{Mode 2 perception system for object detection and color recognition with YOLO v3.}
\label{fig7}
\end{figure}

Transferred to the dialogue interaction, we develop two different HRI styles (Fig. 8) aligning to the combination of color recognition and YOLO v3 object detection. The first case aims to help people get objects with the specific included colors like pink or green, then our “Kinova Gemini” will tell the person all sequencely detected objects with color information one by one and process the YOLO v3 generated “label” by the spaCy NLP library to find if the user-desired color (e.g., Pink, Green) is contained. If true, the robot will ask people “do you want to get this object?” and we can respond with “yes” to grasp it or say other non-yes words to try again for the circular detection and inquiry steps.

\begin{figure}[t]
\centering
\includegraphics[width=1.0\columnwidth]{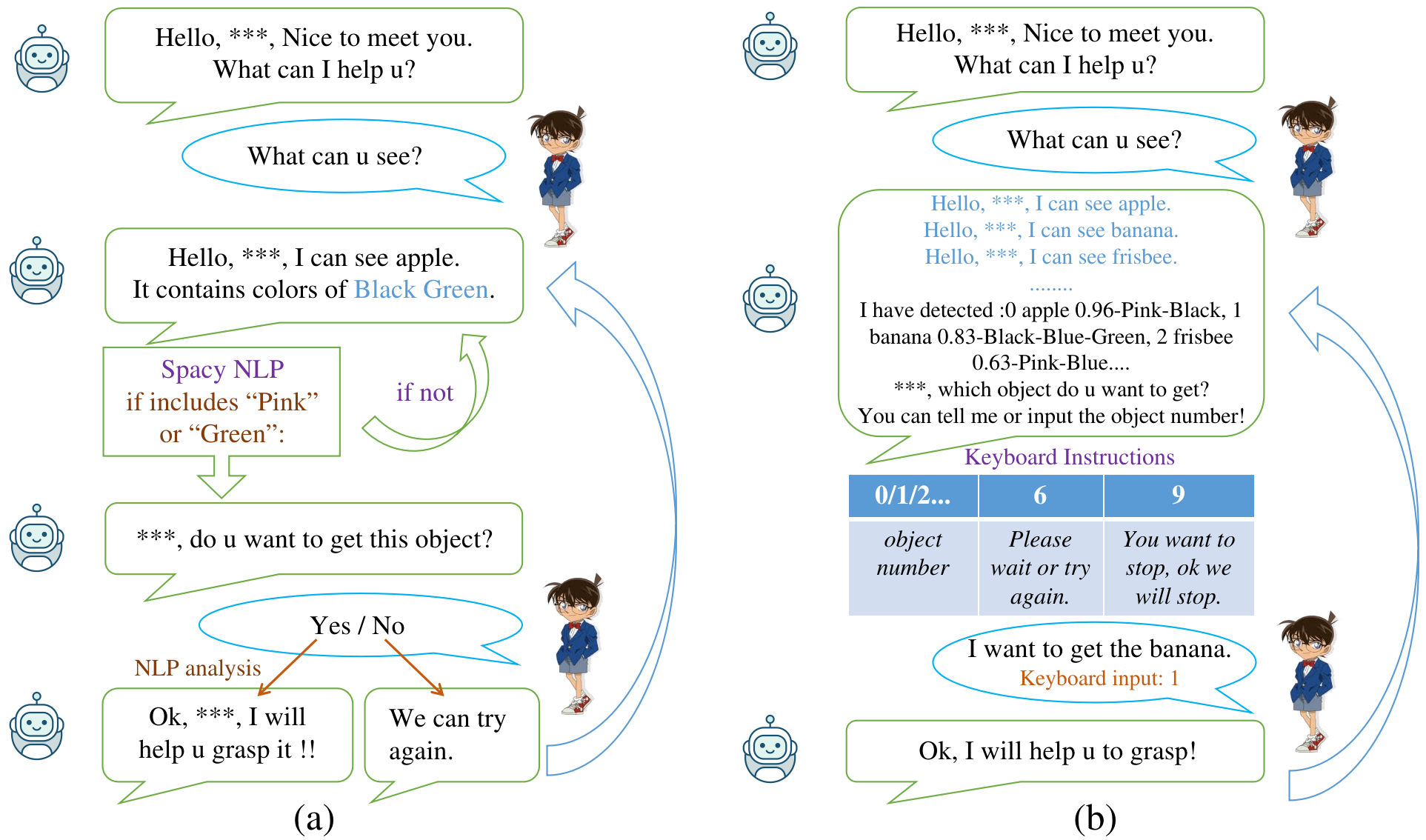}
\caption{Two dialogue interaction styles of Mode 2.}
\label{fig8}
\end{figure}

Consider that the case 1 dialogue style (Fig. 8 (a)) may be a little complicated, so we upgrade the voice system to be much simpler, clearer, and more intelligent as shown in Fig. 8 (b). After greeting with the user, the “Kinova Gemini” can speak out all the detected colorful items at once and inquire people “which object do you want to get?” with the succinct guidance that “You can tell me or input the object number”. Currently, the user can take advantage of quick keyboard interaction as follows: (1) Input the index of detected object that you want to get and make the python script publish corresponding image coordinates to ROS Kortex for grasping; (2) Press “6” to wait or try again, which means that you can wait for much better real-time object detection results and a little delayed visual presentation from OpenCV in the next loop; (3) Input “9” to shut down the whole robot grasping perception program. In this section, the user can continue speaking requests in the dialogue while operating the keyboard input to make the “Kinova Gemini” successfully complete diverse tasks such as “taking multiple objects to the user in the certain order” or “picking up the one with specific color among similar items”.
\begin{figure}[t]
\centering
\includegraphics[width=1.0\columnwidth]{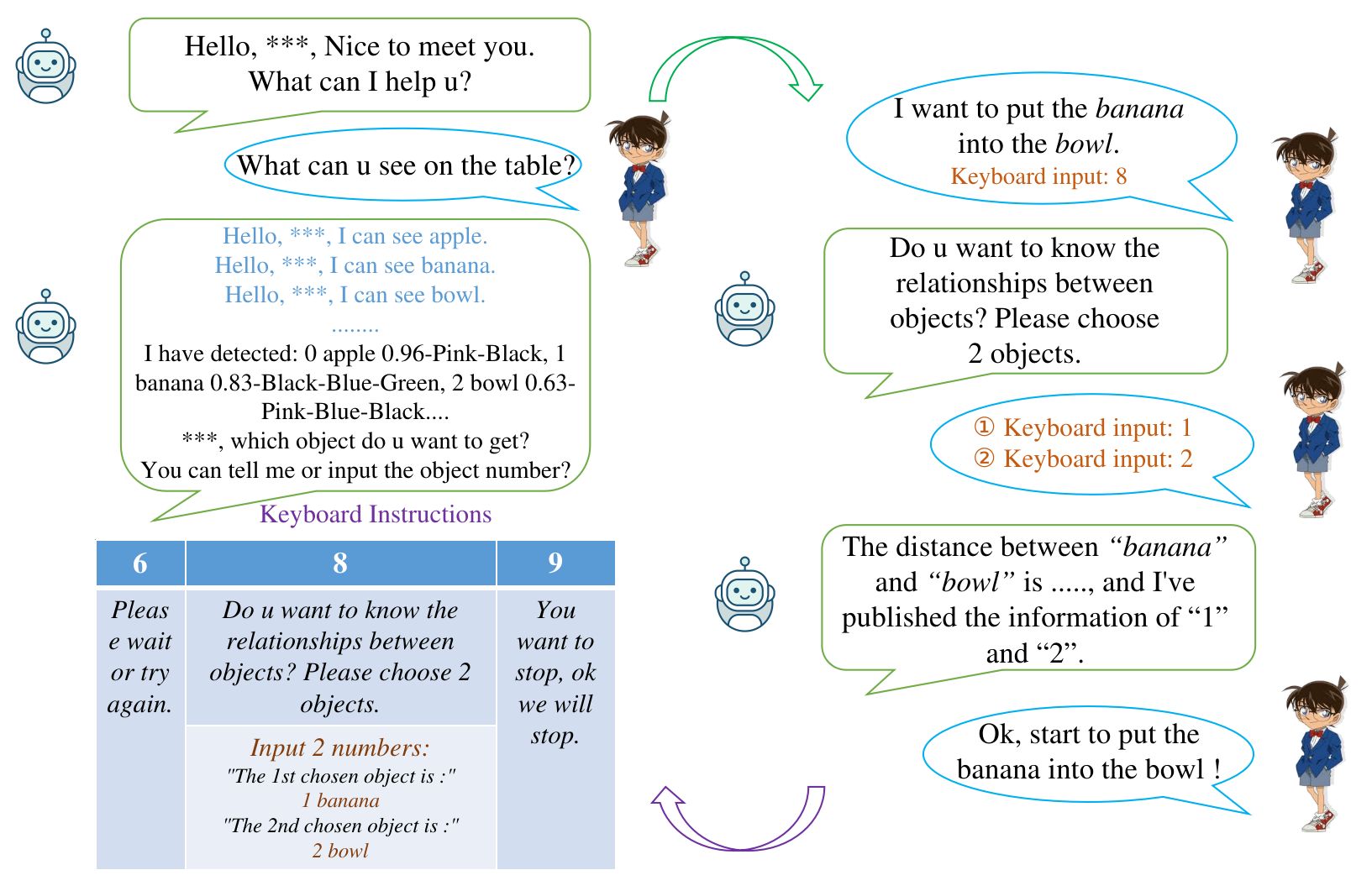}
\caption{The upgraded Mode 3 dialogue interaction.}
\label{fig9}
\end{figure}
\subsection{Mode 3 -- Put the Banana into the Bowl}
Based on previous successful experiments in Mode 1 and Mode 2, we upgrade the “Kinova Gemini” to conduct visual reasoning on deeper analysis of detected objects for a little complicated tasks like “put the banana into the bowl”, which requires detecting multiple objects with YOLO v3 and publish accurate world coordinates of two selected grasp-place objects (e.g., banana-bowl) to ROS Kortex for execution. To achieve this, we add a new keyboard function of “8” in the original Fig. 8 (b) dialogue framework as shown in Fig. 9. When people desire to firstly grasp one detected object and then place it into another detected item, they input “8” after the Kinova robot tells you its YOLO v3 visual perception information for guiding you to choose two objects that the $1^{st}$ item needs to be grasped up and the $2^{nd}$ one means where the $1^{st}$ object will be placed. After that, “Kinova Gemini” can compute the $L2$ distance between the chosen objects and publish their positions to ROS Kortex grasping execution. Therefore, Kinova Gemini can exactly assist the user well to put the banana into the bowl (Fig. 10) with unequivocal steps: (i) The Kinova robot firstly reaches the “grasp” object banana. (ii) Pick the “banana” up; (iii) Place the “banana” into the $2^{nd}$ chosen object “bowl”; (iv) The robot comes back to the “home” state. More experimental results of pick-\&-place tasks (e.g., “put the spoon into the bowl”) can be observed at: \url{https://github.com/2000222/Kinova-Gemini}.

\begin{figure}[t]
\centering
\includegraphics[width=0.8\columnwidth]{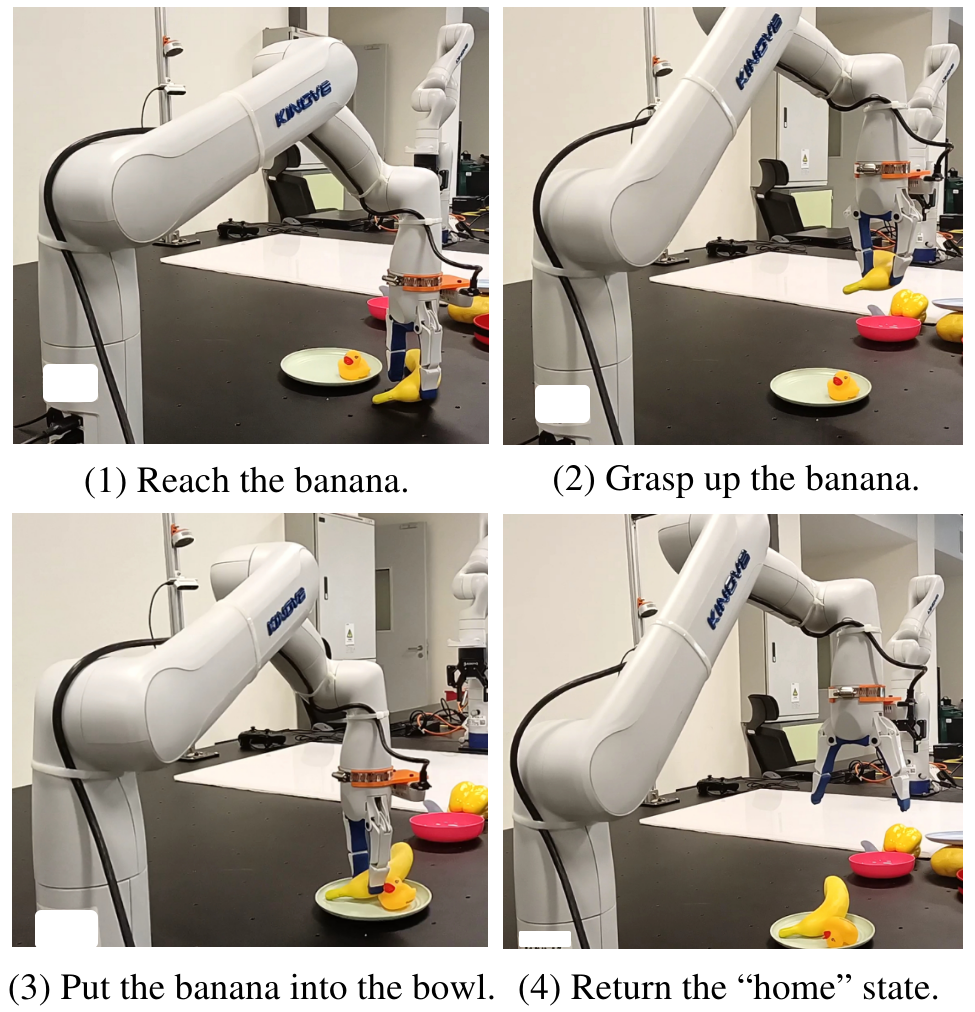}
\caption{“Put the banana into the bowl” task steps.}
\label{fig10}
\end{figure}

\section{Conclusion}
We present the original “Kinova Gemini”, a robotic system that successfully implements the interactive robot grasping or pick-and-place tasks with visual reasoning and conversational AI via HRI in three different application scenarios. Thus, our “Kinova Gemini” can not only assist users to retrieve objects by the GG-CNN auto-grasp detection within dialogue interaction, but also apply the YOLO v3 model for object detection along with color recognition to conduct grasps of specific items, even finish visual reasoning to achieve the complicated pick-and-place tasks (e.g., “put the banana into the bowl”) with human collaboration. All the experimental results of “Kinova Gemini” can refer to \url{https://github.com/2000222/Kinova-Gemini}. In general, “Kinova Gemini” can complete multiple tasks in 3 modes successfully, but the dialogue interaction system with “SpeechRecognition” and “spaCy” can be updated later to be more accurate and effective to recognize human language. Moreover, for future work we may design delicate interaction scenarios with more intelligent conversational AI techniques for “Kinova Gemini” to be used on healthcare, service industry, office, or home.

\end{document}